\documentclass[11pt]{article}

\usepackage[margin=1in]{geometry}
\usepackage[T1]{fontenc}
\usepackage[utf8]{inputenc}
\usepackage{lmodern}
\usepackage{microtype}

\usepackage{amsmath,amssymb}
\usepackage{booktabs}
\usepackage{multirow}
\usepackage{graphicx}
\graphicspath{{figures/}}

\usepackage[numbers,sort&compress]{natbib}
\usepackage{xcolor}
\usepackage[colorlinks=true,linkcolor=blue!60!black,citecolor=blue!60!black,urlcolor=blue!60!black]{hyperref}

\newcommand{\arm}[1]{\texttt{#1}}
\newcommand{\sparseset}{sparse-800}
\newcommand{\oldset}{old-319}

\title{Hints, Critics, and Teachers: Prior Injection for Sparse-Reward RL\\
in Vision--Language Math Reasoning}

\author{Qiqian Fu\\
\texttt{qiqianf2@illinois.edu}}

\date{August 2026}

\begin{document}
\maketitle

\begin{abstract}
Reinforcement learning for vision--language math reasoning starves under sparse
reward: on a pool of 20{,}830 visual-math problems where Qwen2-VL-2B answers
3.6\% of rollouts correctly, 85--97\% of GRPO rollout groups are entirely wrong
and contribute zero gradient. We train eleven methods under identical conditions
in this regime, each injecting a different prior: text (reference-solution
hints), distribution (on-policy distillation from a 7B teacher), and value (a
value-pretrained critic with an MSE or HL-Gauss categorical loss). A prior
helps exactly when it is \emph{delivered}: the six arms whose prior
effectively reaches the policy separate with no overlap from the remaining
five---the no-prior baseline and four arms whose prior is teacher-capped,
gated away, or lost to a mis-parameterized critic---both on the pooled
in-domain metric and on cross-domain transfer (DynaMath). The central
finding, however, concerns evaluation: one slice of the in-domain
pool---long used as this project's general-distribution
check---\emph{anti-correlates} with genuine cross-domain
transfer (Spearman $\rho = -0.74$, $n = 11$ arms, permutation $p = 0.011$),
while the hardest in-domain
slice predicts it closely ($\rho = +0.89$, $p < 0.001$). We attribute the
inversion to a
near-chance multiple-choice subset that rewards models for not having changed;
read through it, the best cross-domain method looked mediocre and the worst
looked like the champion. Among the methods, hint-guided exploration---not
UFT's auxiliary loss---drives hint gains, and replacing the critic's MSE loss
with HL-Gauss cross-entropy is worth $+14.4$ points in-domain. All accuracies
are blind-judged, with paired exact tests.
\end{abstract}

\section{Introduction}
\label{sec:intro}

Group-relative policy optimization (GRPO)~\citep{shao2024deepseekmath} has
become the default recipe for reinforcement learning on math reasoning, but it
carries a structural failure mode: its advantage is computed \emph{within} a
group of rollouts for the same prompt, so a group in which every rollout is
wrong contributes exactly zero gradient. On problems a model rarely solves,
most groups are all-wrong and training starves. This regime is not exotic. When
we profile Qwen2-VL-2B~\citep{wang2024qwen2vl} on the hardest problems of its
own training distribution---a pool of 20{,}830 visual-math problems filtered
from MathV360K~\citep{shi2024mathllava}---the base model answers 3.6\% of
rollouts correctly, and 85--97\% of GRPO groups open all-wrong.

The literature offers three families of remedies, each injecting a different
\emph{prior} into the starved policy. A \textbf{text} prior prefixes rollouts
with part of a reference solution, so the model explores from a state where
success is likely~\citep{liu2025uft,agrawal2026offcontext}. A
\textbf{distribution} prior replaces or augments the reward with a per-token
signal from a stronger teacher, via on-policy
distillation~\citep{lu2025onpolicy,liao2026prefix}. A \textbf{value} prior
trains a critic so that credit reaches individual tokens even when sequence
outcomes are uniform~\citep{yue2025vapo}, and recent work argues the critic's
loss should be categorical rather than a scalar
regression~\citep{imani2018improving,farebrother2024stop,zhou2026start}. These
families are studied in separate papers, on different models, tasks, and
budgets; to our knowledge they have not been compared under one roof, and
almost never on a vision--language model.

This paper runs that comparison. We construct a genuinely sparse training pool
(verifying, before training anything, both that GRPO starves on it and that
hints can unlock it), then train eleven arms---plain GRPO, three hint variants, four
distillation variants, two critic variants, and one arm stacking the text
and distribution priors---under identical conditions: the same pool, the same step and batch budget, the same seed, one
8$\times$A800 node per arm. Every accuracy we report is blind-judged by an
independent LLM judge (DeepSeek)~\citep{deepseekai2024deepseekv3,zheng2023judging}
that reads full outputs without knowing which arm produced them, and pairwise
claims use McNemar exact tests.

The method-level results are clean. What separates the arms is not whether a
prior is injected by design but whether it is \emph{delivered}: on the only
evaluation set outside the training domain (DynaMath~\citep{zou2024dynamath}),
the six arms whose prior demonstrably reaches the policy occupy the range
30.6--32.9\%, and the remaining five---plain GRPO and four arms whose prior
is capped at the teacher's ability, gated to failures, or lost in a
mis-parameterized critic---occupy 27.0--29.1\%, with no overlap. Within that
headline, the
comparison localizes \emph{which} component earns the gain: hint-guided
exploration works while UFT's auxiliary log-likelihood term contributes
nothing; swapping the critic's clipped-MSE loss for HL-Gauss cross-entropy---a
one-line change---moves in-domain accuracy by $+14.4$ pp and turns the weakest
trained arm into the second-best cross-domain arm; annealed global
distillation works where failure-gated persistent distillation does not; and
stacking two priors buys nothing, because both solve the same early bottleneck.

The finding that reorganized this study, however, is about evaluation rather
than methods. For four rounds of this project, a held-out MathV360K slice
(\oldset) served as the ``general-distribution'' check, and conclusions were
drawn from it. With eleven arms evaluated on both domains---and the
in-domain pool decomposed into its two slices---we could finally correlate
the evaluation sets against each other, and \oldset{} ranks the
arms \emph{backwards}: its scores anti-correlate with genuine cross-domain
transfer (Spearman $\rho = -0.74$, $n = 11$ arms), while the hardest
in-domain slice (\sparseset) predicts transfer closely
($\rho = +0.89$). The mechanism, we argue, is compositional: 69\% of \oldset{} is
four-option multiple choice on which every arm sits near the 25\% guessing
floor, so scoring well there means still emitting short option letters like
the base model. The set rewards \emph{not having changed}---precisely the
quantity that anti-correlates with capability gain. Three earlier conclusions
of this project, including an apparent specialization--generalization
trade-off, were artifacts of that lens; we retract them explicitly in
Section~\ref{sec:results}. We report this in detail because the failure
pattern is generic: any project that tracks generalization on a fixed held-out
slice whose composition drifts toward chance-level subtasks can be steered
backwards by it~\citep{alzahrani2024benchmarks}.

A third thread runs through the paper: verification as a deliverable. Earlier
rounds of this project were rewritten by silent scoring and porting bugs, so
round~4 treated every load-bearing mechanism as something to be gated before
training and audited after. A scoring audit found that a heuristic answer
extractor penalized different arms by different amounts---up to 30 pp---in a
direction that inverted rankings; a hacking analysis tested and rejected
guessing and format-reward explanations for the headline gains
\citep{shao2025spurious}. We describe this discipline in
Section~\ref{sec:verification} because without it, most of the numbers in this
paper would have been wrong in ways no significance test would catch.

\paragraph{Contributions.}
\begin{itemize}
  \item A controlled comparison of eleven training arms spanning text,
    distribution, and value priors for sparse-reward RL on a 2B
    vision--language model, under identical data, budget, and judging, with
    verified preconditions of sparsity (85--97\% all-wrong groups) and hint
    efficacy (7.2\% $\to$ 34.4\% with a half-solution prefix).
  \item Method findings: arms with a delivered prior separate cleanly from
    the rest on cross-domain transfer; exploration, not the auxiliary objective,
    drives hint gains; the critic's loss parameterization alone is worth
    $+14.4$ pp (concurrent with a text-only version of this finding
    \citep{zhou2026start}); two priors do not compose.
  \item An evaluation-validity finding: a held-out ``general'' set that
    anti-predicts cross-domain transfer ($\rho = -0.74$) while the hardest
    in-domain tail predicts it ($\rho = +0.89$), a mechanism for the
    inversion, and three explicit retractions of conclusions the broken lens
    had produced.
  \item A reusable verification protocol for RL comparisons---blind LLM
    judging, scoring audits, pre-training hard gates, and reward-hacking
    analyses---together with the concrete failures it caught.
\end{itemize}

\paragraph{Organization.}
Section~\ref{sec:related} situates the three prior families.
Section~\ref{sec:setup} describes the pool construction (including a
measurement trap that would have selected formatting failures instead of hard
problems), the eleven arms, and the judging protocol.
Section~\ref{sec:results} presents the main comparison and the evaluation-set
inversion. Section~\ref{sec:findings} details seven method-level findings.
Section~\ref{sec:verification} documents the verification discipline, and
Section~\ref{sec:limitations} states limitations, chief among them the single
training seed.

\section{Related Work}
\label{sec:related}

\paragraph{Guided exploration with hints.}
UFT \citep{liu2025uft} unifies supervised and reinforcement fine-tuning by prefixing rollouts with part of a reference solution, annealing the hint proportion toward zero over training, and adding a supervised log-likelihood term on hint tokens to the RL objective. The hint acts as an exploration scaffold: it places the policy in the neighborhood of a correct solution, where reward becomes reachable. Our \arm{hintonly}, \arm{uft001}, and \arm{uft002} arms instantiate this recipe on top of GRPO and deliberately factor it into its two components---hint-guided exploration and the auxiliary likelihood term---so that each component can be credited separately.

\paragraph{On-policy distillation.}
On-policy distillation trains the student on its own rollouts while a teacher scores every token, setting the per-token advantage to the negative reverse KL between student and teacher \citep{lu2025onpolicy}. \citet{liao2026prefix} extend this to multi-turn agents by replaying pre-collected teacher trajectories as rollout prefixes under a step-decaying length schedule (ReOPD); our \arm{reopd} arm adapts the replay mechanism to our single-turn setting. Because reverse-KL imitation cannot exceed the behavior it imitates, distillation-only training inherits the teacher's ceiling---a property our comparison makes visible by running distillation arms alongside reward-driven arms that are free to surpass the teacher.

\paragraph{Value functions and critic parameterization.}
Classic actor--critic training pairs PPO \citep{schulman2017proximal} with generalized advantage estimation \citep{schulman2015high}. VAPO \citep{yue2025vapo} adapts value-based RL to long-chain-of-thought reasoning through value pretraining and decoupled GAE, and adopts the clip-higher technique introduced by DAPO \citep{yu2025dapo}; our \arm{vapo} arm follows this recipe. A separate line of work replaces the critic's scalar regression with classification over a discretized return support: the histogram loss of \citet{imani2018improving}, which \citet{farebrother2024stop} showed to scale value learning across deep-RL domains, and which \citet{zhou2026start} applied to LLM RL critics---concurrently with our work, and on text-only tasks. Our \arm{vapohl} arm isolates exactly this substitution in a vision--language setting.

\paragraph{RL for vision--language math reasoning.}
GRPO \citep{shao2024deepseekmath} has become the default policy-gradient method for math reasoning and transfers directly to vision--language models such as Qwen2-VL \citep{wang2024qwen2vl}. Training corpora and benchmarks for this setting include MathV360K \citep{shi2024mathllava} and the program-generated DynaMath \citep{zou2024dynamath}. Several works caution against taking RLVR gains at face value: accuracy-only rewards fail to improve the perception component of multimodal reasoning \citep{xiao2025perception}, even spurious rewards can raise benchmark scores \citep{shao2025spurious}, and GRPO's normalization terms carry biases of their own \citep{liu2025understanding}, which in all-failure groups can let auxiliary dense terms dominate the gradient \citep{wang2026dark}. Closest to our hint arms, Off-Context GRPO \citep{agrawal2026offcontext} also conditions rollouts on privileged information but corrects the resulting off-policyness with an importance-weighted objective, a correction our hint arms do not apply.

\paragraph{Evaluation validity.}
All accuracies we report are blind-judged by an LLM judge, a protocol whose agreement with human raters is well documented \citep{zheng2023judging}, as are its failure modes \citep{gu2024survey}. On the benchmark side, \citet{alzahrani2024benchmarks} show that the surface composition of multiple-choice benchmarks---option order, answer-extraction format---can move leaderboard rankings by several positions: what a benchmark appears to measure and what actually drives its rankings can diverge. Our evaluation-validity finding is of this kind but sharper in direction---a held-out set used across every round of this project as its generalization check turns out to rank methods in the \emph{reverse} order of their true cross-domain transfer.

\section{Experimental Setup}
\label{sec:setup}

\subsection{Background: three rounds of honest negatives}
\label{sec:background}

This work is the fourth round of a longer investigation. Rounds~1--3 asked
whether UFT-style hint-guided exploration \citep{liu2025uft} beats plain GRPO
\citep{shao2024deepseekmath} for a 2B vision--language model on visual math,
and concluded---after retracting two rounds contaminated by five port-fidelity
bugs (whose fixes moved individual arm scores by up to $+9.1$ pp)---that
\emph{UFT ties GRPO}: in round~3, \arm{grpo} reached 33.9\% on the held-out
set against 31.0/32.6\% for the two UFT arms, with no significant differences
in the hint arms' favor. The mechanism analysis of the earlier rounds contained the
seed of this one: the reward was never sparse enough for hints to matter.
GRPO's fraction of all-wrong rollout groups---groups that contribute zero
gradient under group-normalized advantages---opened at 77\% and fell to
roughly 30\% within training; the hint had nothing to rescue.

Round~4 tests the counterfactual: \emph{make the reward genuinely sparse,
then re-run the comparison}---and widens it to every prior-injection family
proposed for that regime.

\subsection{Constructing a genuinely sparse pool}
\label{sec:pool}

We profiled Qwen2-VL-2B-Instruct \citep{wang2024qwen2vl} on all 52{,}964
problems of our MathV360K-derived training corpus \citep{shi2024mathllava}:
8 rollouts per problem at temperature 1.0, roughly 428k generations in total.
Profiling immediately surfaced a measurement trap worth naming.

\begin{quote}
\textbf{Measurement trap.} Scored with the training reward---which requires
an \texttt{Answer: X} format---the base model gets \textbf{3.5\%}. Rescored
semantically, the same outputs score \textbf{43.4\%}. The strict scorer
measures format compliance, not ability; selecting ``hard'' problems with it
would have built a pool of formatting failures. All difficulty selection
therefore used the semantic score, and every reported result uses a blind
LLM judge that reads full outputs and is immune to format drift.
\end{quote}

The pool consists of the 21{,}630 problems the base model solves at most
1-in-8 times under semantic scoring. We hold out an 800-problem split
(\sparseset{}) and train on the remaining 20{,}830. Within the pool the base
model answers 3.6\% of rollouts correctly.

Two preconditions were verified \emph{before} training a single arm. First,
the pool is genuinely sparse for GRPO: the opening all-wrong group fraction
is 85--97\%, versus 77\% in the earlier rounds. Second, hints can actually unlock it:
prefixing half a reference solution lifts base accuracy from 7.2\% to
34.4\%---a property round~3's pool never had.

\subsection{The eleven arms}
\label{sec:arms}

Eleven methods were trained under identical conditions, each injecting a
different prior into the starved regime (Table~\ref{tab:mechanisms}).

\begin{table}[t]
\centering
\small
\begin{tabular}{lll}
\toprule
Arm & Prior & Mechanism \\
\midrule
\arm{grpo}     & none         & plain GRPO, groups of $n{=}4$ (the baseline) \\
\arm{hintonly} & text         & reference-solution prefix, cosine-annealed $95\%{\to}5\%$, then off \\
\arm{uft001}   & text         & \arm{hintonly} + NLL on hint tokens, coef.\ 0.001 (full UFT) \\
\arm{uft002}   & text         & \arm{hintonly} + NLL on hint tokens, coef.\ 0.002 \\
\arm{opd}      & distribution & pure on-policy distillation: advantage $=-$reverse KL, no reward \\
\arm{reopd}    & distribution & \arm{opd} + teacher-trajectory prefix replay \\
\arm{srpo}     & distribution & GRPO + distillation gated to failed rollouts only \\
\arm{saf}      & distribution & GRPO + distillation clamped to $\pm 2$, annealed $\lambda(t){:}\,1{\to}0$ \\
\arm{hintsaf}  & text + dist. & \arm{saf} + hint injection \\
\arm{vapo}     & value        & PPO with value-pretrained critic, clipped-MSE value loss \\
\arm{vapohl}   & value        & \arm{vapo} with HL-Gauss categorical value loss \\
\bottomrule
\end{tabular}
\caption{The eleven arms. All train on the same 20{,}830-problem sparse pool
for 500 steps with identical batch size, learning rate, and seed.}
\label{tab:mechanisms}
\end{table}

\paragraph{Text priors.}
\arm{hintonly} prefixes each rollout with part of a reference solution; the
fraction of hinted rollouts is cosine-annealed from 95\% to 5\% over the
first 300 steps, then set to zero. \arm{uft001} and \arm{uft002} add the
signature term of UFT \citep{liu2025uft}---a log-likelihood loss on the hint
tokens---at coefficients 0.001 and 0.002, completing the full UFT objective.

\paragraph{Distribution priors.}
\arm{opd} is pure on-policy distillation from a Qwen2-VL-7B teacher in the
Thinking Machines recipe \citep{lu2025onpolicy}: the per-token advantage is
the negative reverse KL to the teacher (k1 estimator), with no task reward.
\arm{reopd} adds teacher-trajectory prefix replay with a geometrically
sampled cut index ($\kappa=0.6$), our single-turn adaptation of a method
originally proposed for multi-turn agents \citep{liao2026prefix}.
\arm{srpo} keeps the full GRPO objective and gates the distillation signal
to failed rollouts only (sequence advantage $\le 0$), never decaying it.
\arm{saf} instead applies the distillation term globally, clamped to $\pm 2$
and cosine-annealed from 1 to 0 over the 500 steps. \arm{hintsaf} stacks
\arm{saf} with hint injection to test whether the two priors compose.

\paragraph{Value priors.}
\arm{vapo} follows the VAPO recipe \citep{yue2025vapo}: PPO
\citep{schulman2017proximal} with a same-backbone critic, 30 steps of value
pretraining against Monte-Carlo targets with the actor frozen, decoupled GAE
\citep{schulman2015high} with $\lambda_{\text{critic}}=1$, and the
clip-higher ratio bounds $0.2/0.28$ introduced by DAPO \citep{yu2025dapo}
and adopted by VAPO. We set $\lambda_{\text{policy}}=0.98$ (our choice; VAPO
proposes a length-adaptive schedule). The value loss is clipped MSE.
\arm{vapohl} changes exactly one thing: the value loss becomes an HL-Gauss
cross-entropy \citep{imani2018improving,farebrother2024stop} over 101 bins
on $[-0.1, 1.1]$ with $\sigma = 0.75\times$ bin width, truncated-Gaussian
targets, no value clipping, and a zero-initialized head.

\subsection{Training configuration}
\label{sec:training}

Every arm trains for 500 steps at batch size 256 prompts $\times$ 4 rollouts,
learning rate $10^{-6}$, KL coefficient 0.001, seed 42, on one
8$\times$A800 node (distillation arms split the node 4/4 between student and
teacher). The teacher's tokenizer is file-identical to the student's
(MD5-verified; Section~\ref{sec:verification}). All training runs on verl
\citep{sheng2024hybridflow} with vLLM rollouts \citep{kwon2023efficient};
our modifications to the framework are environment-gated patches described
in Section~\ref{sec:verification}.

\subsection{Evaluation sets and judging protocol}
\label{sec:evalsets}

We evaluate along two axes (Table~\ref{tab:evalsets}): one in-domain
dataset drawn from MathV360K, and one cross-domain dataset with no overlap
with it. The in-domain dataset (MathV, $n{=}1119$) pools two slices at
their size weights (71.5/28.5): \sparseset{}, the held-out hard tail of the
training distribution, and \oldset{}, a different MathV360K slice (rows
excluded from training by an earlier data-cleaning step), 69\% of which is
Geometry3K four-option multiple choice and which had served as this
project's general-distribution check since round~1. The cross-domain
dataset is DynaMath \citep{zou2024dynamath} ($n{=}477$, an easy split of
373 problems and a hard split of 104): its problems are programmatically
rendered, with zero overlap with MathV360K. Because the two in-domain
slices turn out to measure different quantities
(Section~\ref{sec:results-inversion}), we report the pooled in-domain
number together with its per-slice breakdown throughout.

\begin{table}[t]
\centering
\small
\begin{tabular}{llrlr}
\toprule
Domain & Set & $n$ & Nature & Base (\%) \\
\midrule
in-domain & MathV (pooled) & 1119 & \sparseset{} $+$ \oldset{}, size-weighted 71.5/28.5 & \\
& \quad \sparseset{}  & 800 & training-distribution hard tail, never trained on & $\approx 0$ \\
& \quad \oldset{}     & 319 & earlier MathV360K slice; 69\% Geometry3K 4-option MC & --- \\
\midrule
cross-domain & DynaMath & 477 & programmatically rendered; zero overlap with MathV360K & 25.8 \\
& \quad easy & 373 & & 30.3 \\
& \quad hard & 104 & & 11.5 \\
\bottomrule
\end{tabular}
\caption{Evaluation datasets, grouped by domain. The combined DynaMath base
figure is reported as measured; the size-weighted average of the two split
figures is 26.2, a 0.4~pp bookkeeping difference we do not resolve here.}
\label{tab:evalsets}
\end{table}

\paragraph{Judging.}
All reported accuracies come from a model-blind LLM judge
(DeepSeek; \citealp{deepseekai2024deepseekv3}) that reads the full model
output with the arm identity hidden and judges each answer against the
ground truth. Pairwise comparisons use McNemar's exact test on paired
per-problem outcomes. Judge run-to-run variance is small: three independent
judging passes over identical \arm{grpo} outputs scored 40.5, 41.1, and
41.0 ($\pm 0.5$ pp).

\section{Results}
\label{sec:results}

\subsection{Main comparison}
\label{sec:results-main}

Table~\ref{tab:main} reports blind-judged accuracy for all eleven arms on
the in-domain metric (MathV pooled, with its two-slice breakdown) and the
cross-domain metric (DynaMath), sorted by DynaMath. Every trained arm
clears the near-zero base accuracy on the training distribution's hard tail
by a wide margin, but the arms divide sharply in how far they get: six arms
span 41.6--44.1 on the in-domain metric (49.4--53.6\% on the \sparseset{}
slice), while the other five span 34.9--37.8 (38.8--41.0 on \sparseset{}).
The same six-versus-five split appears on both axes.

We refer to the upper six as the \emph{delivered-prior} arms and the lower
five as the \emph{undelivered-prior} arms. The names are empirical rather
than architectural---\arm{opd}, \arm{reopd}, \arm{srpo}, and \arm{vapo} all
inject a prior by design---but every membership is explained by a design
feature: the lower group comprises the no-prior baseline (\arm{grpo}), pure
distillation capped at its teacher's own ability (\arm{opd}, \arm{reopd}),
distillation gated to failed rollouts and never decayed (\arm{srpo}), and a
value prior lost to a mis-parameterized critic (\arm{vapo};
Section~\ref{sec:findings}).

\begin{table}[t]
\centering
\caption{Blind-judged accuracy (\%) of all eleven arms, sorted by the
cross-domain metric (DynaMath, n=477, pooling the easy and hard splits).
MathV (pooled, n=1119) is the in-domain metric, combining the \sparseset{}
and \oldset{} slices at their size weights (71.5/28.5); the two slice
columns give its breakdown. The horizontal rule separates the six
delivered-prior arms from the five undelivered-prior arms: the two groups
do not overlap on either metric. Within-group orderings should not be read
as rankings (single training seed; Section~\ref{sec:limitations}).}
\label{tab:main}
\begin{tabular}{llcccc}
\toprule
Arm & Prior & \sparseset{} & \oldset{} & MathV (pooled) & DynaMath \\
\midrule
\arm{uft002}   & text                & 52.4 & 19.1 & 42.9 & \textbf{32.9} \\
\arm{vapohl}   & value (HL-Gauss)    & \textbf{53.6} & 19.7 & 44.0 & 31.9 \\
\arm{hintonly} & text                & 51.4 & 21.3 & 42.8 & 31.2 \\
\arm{hintsaf}  & text + distribution & 49.4 & 26.0 & 42.7 & 30.8 \\
\arm{uft001}   & text                & 52.0 & 15.7 & 41.6 & 30.8 \\
\arm{saf}      & distribution (annealed) & 49.6 & 30.4 & \textbf{44.1} & 30.6 \\
\midrule
\arm{srpo}     & distribution (gated) & 40.8 & 30.4 & 37.8 & 29.1 \\
\arm{opd}      & distribution (pure)  & 40.5 & 22.9 & 35.5 & 29.1 \\
\arm{reopd}    & distribution (replay) & 38.8 & 25.1 & 34.9 & 28.1 \\
\arm{grpo}     & none                & 41.0 & 27.6 & 37.2 & 27.5 \\
\arm{vapo}     & value (MSE)         & 39.2 & \textbf{32.6} & 37.4 & 27.0 \\
\midrule
base (untrained) & ---               & $\sim$0 & --- & --- & 25.8 \\
\bottomrule
\end{tabular}
\end{table}

\begin{figure}[t]
\centering
\includegraphics[width=0.85\linewidth]{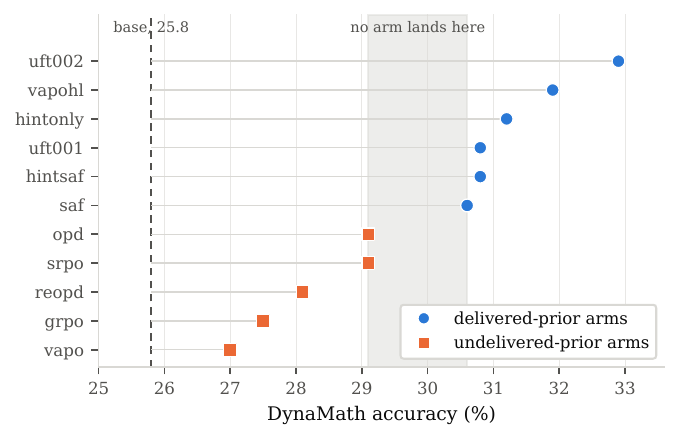}
\caption{All eleven arms ranked by DynaMath accuracy. Every delivered-prior
arm outperforms every undelivered-prior arm: the shaded band (29.1--30.6) separates
the two groups and contains no arm. Differences within each group are inside
judging noise and should not be ranked (Section~\ref{sec:limitations}).
Dashed line: untrained base (25.8).}
\label{fig:arms}
\end{figure}

The cross-domain column exhibits a complete separation
(Figure~\ref{fig:arms}): the six delivered-prior arms occupy every position
from 30.6 to 32.9, and the five undelivered-prior arms every position from 27.0 to
29.1, with no overlap between the groups. We emphasize what this claim is
and is not: the \emph{between-group} separation is the result; the ordering
\emph{within} either group spans at most 2.3 pp on a 477-item set under a
single training seed, and we do not interpret it
(Section~\ref{sec:limitations}).

The group boundary also clarifies why \arm{vapo} and \arm{vapohl}---which
share every component except the critic's loss function---land on opposite
sides. A value prior is only injected to the extent that the critic actually
learns value; with a clipped-MSE loss the critic starts from a pathological
cold state and never catches up, so \arm{vapo} behaves like an arm without a
prior, while the HL-Gauss parameterization delivers the prior and places
\arm{vapohl} with the delivered group (Section~\ref{sec:findings}, F3).

\subsection{The benchmark that ranked backwards}
\label{sec:results-inversion}

The finding that reorganized this report concerns the evaluation rather
than the methods: the single in-domain number of Table~\ref{tab:main} hides
two components that measure different quantities. Since round~1 this
project had used \oldset{}---the slice that now forms 28.5\% of the
in-domain pool---as its ``general-distribution'' check. Read against the
cross-domain column, that slice runs the wrong way: \arm{vapo} tops \oldset{}
at 32.6 while placing last among all trained arms on DynaMath (27.0, below
\arm{grpo}), and \arm{vapohl} sits third-from-bottom on \oldset{} (19.7)
while placing second overall on DynaMath (31.9).

\begin{table}[t]
\centering
\caption{Spearman rank correlation between evaluation sets over the eleven
trained arms (n = 11). The five correlations computable from
Table~\ref{tab:main} were independently recomputed from the published
numbers and agree with the values below to within 0.004; the two easy-split
rows use per-arm accuracies on DynaMath-easy alone.}
\label{tab:corr}
\begin{tabular}{llr}
\toprule
X & Y & $\rho$ \\
\midrule
\sparseset{} & DynaMath & $+0.89$ \\
\oldset{}    & DynaMath & $-0.74$ \\
MathV pooled (71.5/28.5) & DynaMath & $+0.76$ \\
MathV pooled (50/50)     & DynaMath & $+0.40$ \\
\sparseset{} & \oldset{} & $-0.66$ \\
\sparseset{} & DynaMath-easy & $+0.87$ \\
\oldset{}    & DynaMath-easy & $-0.81$ \\
\bottomrule
\end{tabular}
\end{table}

\begin{figure}[t]
\centering
\includegraphics[width=\linewidth]{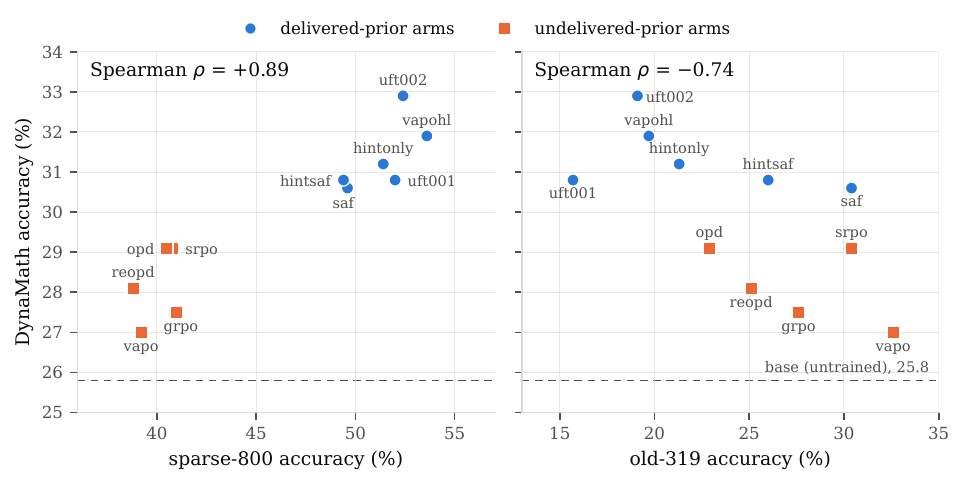}
\caption{Which held-out set predicts cross-domain transfer? Each point is one
trained arm; the $y$ axis is accuracy on DynaMath ($n{=}477$), the only
evaluation pool with no overlap with the training distribution.
\textbf{Left:} the hardest in-domain held-out set (\sparseset{}) ranks arms
nearly as DynaMath does (Spearman $\rho = +0.89$, $n = 11$).
\textbf{Right:} the set used as a ``general-distribution'' check since
round~1 (\oldset{}) anti-predicts transfer ($\rho = -0.74$): \arm{vapo} tops
\oldset{} while placing last among trained arms on DynaMath, and
\arm{vapohl} nearly inverts that. Dashed line: untrained base model (25.8).}
\label{fig:scatter}
\end{figure}

\begin{figure}[t]
\centering
\includegraphics[width=0.7\linewidth]{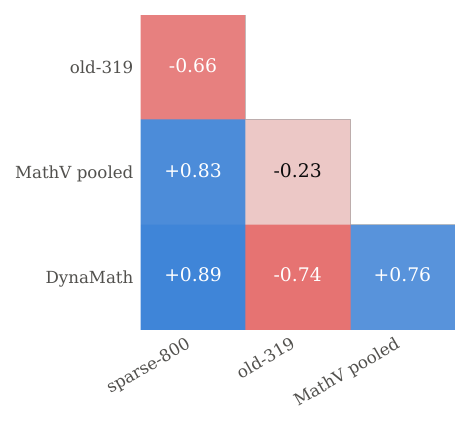}
\caption{Spearman rank correlations between evaluation sets over the eleven
trained arms, computed from the blind-judged accuracies in
Table~\ref{tab:main}. \sparseset{} is the strongest predictor of
cross-domain accuracy ($+0.89$); \oldset{} is strongly anti-correlated with
it ($-0.74$) and with \sparseset{} itself ($-0.66$). The pooled in-domain
metric inherits its predictive validity mechanically from its \sparseset{}
component (the pool contains both sets; weights 71.5/28.5).}
\label{fig:corrmat}
\end{figure}

Table~\ref{tab:corr} and Figures~\ref{fig:scatter} and~\ref{fig:corrmat}
quantify the pattern. The hardest in-domain set, \sparseset{}, predicts
genuine cross-domain transfer closely ($\rho = +0.89$, n = 11
arms, permutation $p < 0.001$), while \oldset{} \emph{anti-predicts} it
($\rho = -0.74$, permutation $p = 0.011$). The two
in-domain sets anti-correlate with each other ($\rho = -0.66$): they are not
noisy measurements of one quantity but measurements of two different
quantities. Because the within-group orderings are inside noise
(Section~\ref{sec:limitations}), these correlations are driven by the
between-group separation and should be read as statements about which sets
order the two groups correctly, not as evidence about fine-grained rankings
of individual arms.

\paragraph{Why \oldset{} inverts.}
69\% of \oldset{} is Geometry3K four-option multiple choice, on which every
arm---trained or untrained---sits near the 25\% guessing floor. Scoring well
there is therefore not about solving geometry; it is about still behaving
like the base model, emitting a short bare option letter. Arms that
genuinely learned something shifted their output distribution toward
multi-step reasoning, which on a chance-level multiple-choice set converts
lucky guesses into reasoned-but-wrong commitments. On this reading, the set
rewards \emph{not having changed}, and that is precisely the quantity that
anti-correlates with capability gain. We state the epistemic status of this
explanation plainly: it is an interpretation consistent with the set's
composition, not a direct multiple-choice versus free-form decomposition of
each arm's predictions, which we defer to a future revision.

\paragraph{Consequences for pooled metrics.}
A corollary is that any single in-domain number inherits the flaw in
proportion to how much \oldset{} it contains: at the 71.5/28.5 size
weighting the pooled metric correlates with DynaMath at $\rho = +0.76$,
but at equal weighting only $+0.40$, because equal weighting amplifies the
anti-correlated half. The cleanest single in-domain number is \sparseset{}
alone.

\paragraph{Corrected readings.}
Three interim conclusions of this project were artifacts of reading
\oldset{} as a generalization measure. We retain them here, with their
corrections, as a record of how a miscalibrated held-out set distorts
method comparison (Table~\ref{tab:corrected}).

\begin{table}[t]
\centering
\caption{Earlier readings and their corrections after the evaluation-validity
analysis.}
\label{tab:corrected}
\begin{tabular}{p{0.44\linewidth}p{0.50\linewidth}}
\toprule
Earlier reading & Corrected \\
\midrule
All arms lie on one specialization-versus-generalization trade-off curve;
none advances both ends. &
The apparent curve is capacity reallocation \emph{within} MathV360K. Across
domains there is no trade-off: stronger in-domain predicts stronger
out-of-domain. \\
\addlinespace
\arm{vapohl} is the specialization extreme, paying a generalization tax. &
\arm{vapohl} is second-best cross-domain (31.9). Its low \oldset{} score
reflects only that it stopped guessing option letters. \\
\addlinespace
\arm{vapo} is the robustness champion---best on the ``general-distribution''
set. &
\arm{vapo} is the weakest trained arm cross-domain (27.0, below plain
\arm{grpo}). \oldset{} had the ranking inverted. \\
\bottomrule
\end{tabular}
\end{table}

\section{Method Findings}
\label{sec:findings}

The seven findings below are stated against the numbers in
Table~\ref{tab:main}; all pairwise claims use exact McNemar tests on
blind-judged outputs.

\paragraph{F1 --- Sparse reward is the regime where hints earn their keep.}
All three hint arms beat \arm{grpo} on \sparseset{} by $+10.4$ to $+11.4$ pp,
every comparison $p < 10^{-4}$ with confidence intervals bounded away from
zero. Round~3 of this project had concluded that UFT ties GRPO
\citep{liu2025uft}; that conclusion is thereby relocated, not refuted. In
the earlier rounds the reward was never sparse enough for the mechanism to
bind---the fraction of all-wrong rollout groups started at roughly 77\% and
fell to about 30\% over training, so the hint had little to rescue. On the present pool, which
opens at 85--97\% all-wrong groups, the same mechanism is worth ten points.

\paragraph{F2 --- It is the exploration, not the objective.}
\arm{hintonly}, \arm{uft001}, and \arm{uft002} are statistically
indistinguishable in-distribution (51.4 / 52.0 / 52.4 on \sparseset{}, all
pairwise n.s.), so UFT's auxiliary hint-token log-likelihood term contributes
nothing where it was supposed to help. The best UFT
configuration is the one with UFT's own signature term removed; what remains
is hint-guided exploration.

\begin{figure}[t]
\centering
\includegraphics[width=0.72\linewidth]{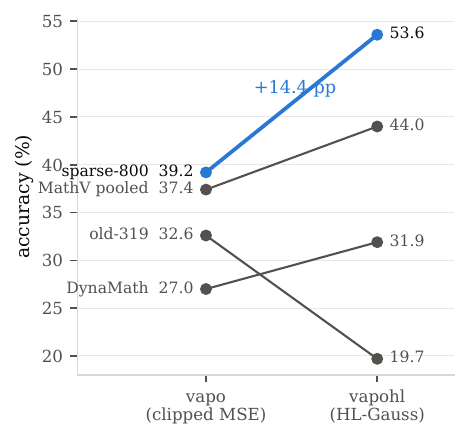}
\caption{One change: replacing \arm{vapo}'s clipped-MSE value loss with
HL-Gauss cross-entropy (\arm{vapohl}), everything else identical. In-domain
accuracy on \sparseset{} moves $+14.4$ pp and cross-domain DynaMath from
27.0 to 31.9, while \oldset{} drops---the arm stopped guessing option
letters (Section~\ref{sec:results-inversion}).}
\label{fig:vapohl}
\end{figure}

\paragraph{F3 --- The critic's loss function is worth more than the critic.}
\arm{vapo} $\to$ \arm{vapohl} changes exactly one component---clipped MSE
becomes HL-Gauss cross-entropy \citep{imani2018improving,farebrother2024stop}
over 101 bins---and moves \sparseset{} from 39.2 to 53.6 ($+14.4$ pp,
significant), DynaMath from 27.0 to 31.9, and explained variance from 0.59
to 0.66 (Figure~\ref{fig:vapohl}). The categorical head also removes a
cold-start pathology: zero-initialized, it starts at a uniform distribution
($V = 0.5$, explained variance $\approx 0$), where the scalar head starts at
an explained variance of $-755$. A good critic decides how deep the policy
can dig; a badly parameterized one wastes the dig. A concurrent study
reports the same categorical-critic effect for text-only LLM reinforcement
learning \citep{zhou2026start}; the two results were obtained independently
and corroborate each other across modalities.

\paragraph{F4 --- Annealed distillation is the robust recipe; targeted
distillation is not.}
\arm{saf}---GRPO plus a bounded, cosine-decayed teacher-KL term---beats
\arm{grpo} by $+8.6$ pp on \sparseset{} and posts the best DynaMath-hard
score of all eleven arms (18.3\%). \arm{srpo} applies the same distillation
signal but gates it to failed rollouts and never decays it, and lands at
\arm{grpo} level on every set. A global prior that fades as the policy finds
its own signal works; a targeted, persistent one does not.

\paragraph{F5 --- Pure distillation inherits its teacher's ceiling.}
\arm{opd} and \arm{reopd} \citep{lu2025onpolicy,liao2026prefix} converge to
\arm{grpo}-level performance. The 7B teacher itself scores only 27.9\% on
the pool; reverse-KL imitation cannot exceed what the teacher knows, while
reward-driven arms can and do---the hint arms reach roughly 52\% on
\sparseset{}, far beyond their teacher. \arm{reopd} tracks \arm{opd}
everywhere (all n.s.), consistent with its source setting: prefix replay
repairs a multi-turn pathology that this single-turn setting does not have.

\paragraph{F6 --- The gains travel.}
Every delivered-prior arm beats the untrained base on DynaMath by $+4.8$ to
$+7.1$ pp, while plain \arm{grpo} manages $+1.7$. On the easy split the hint
arms' margins over base are individually significant ($+6.2$ to $+7.8$ pp,
$p = 0.012$--$0.035$), where \arm{grpo} moves $-0.5$ pp (n.s.). This is the
first significant cross-domain transfer in four rounds of this project, and
it belongs to the priors, not to reinforcement learning as such.

\paragraph{F7 --- Two priors do not compose.}
\arm{hintsaf} stacks the text prior on the annealed distribution prior and
lands at the weighted middle of its parents on every set (\sparseset{} 49.4:
$-0.2$ pp vs.\ \arm{saf} and $-2.0$ pp vs.\ \arm{hintonly}, both n.s.).
Hints and teacher-KL solve the same early bottleneck---escaping the
zero-gradient region of an all-wrong batch---so stacking them is redundant
rather than complementary.

\section{Verification Discipline}
\label{sec:verification}

Rounds~1--2 of this project were retracted after five silent port-fidelity
bugs rewrote their conclusions, so round~4 treated verification as a
deliverable in its own right. Everything in this section happened
\emph{before} conclusions were drawn.

\subsection{The scoring audit that rewrote the leaderboard}
\label{sec:audit}

Before adopting any accuracy number we audited the heuristic
(pattern-extraction) scorer against roughly 340 hand-read rows, backed by a
programmatic recheck of the full result set. The audit found that the
heuristic penalized each arm differently---and in the \emph{opposite}
direction from round~3, because trained arms drift away from the extraction
patterns rather than toward them. \arm{uft001}'s apparent collapse on
\oldset{} (6.9\%) was entirely a parser bug: the arm writes
``\texttt{Final answer: A. 5.}'' and the extractor failed all 141
multiple-choice rows it touched. The distillation arms' terse answers were
under-scored by 19--30 pp. Corrected estimates predicted the subsequent
blind-judge results within about 1 pp---but raw heuristic rankings and
blind-judged rankings disagreed on nearly every pairwise comparison. Only
blind-judged numbers appear in this paper.

\subsection{The hacking analysis}
\label{sec:hacking}

The hint arms' large \sparseset{} gains invited suspicion, so four
reward-hacking hypotheses were tested against them
(Table~\ref{tab:hacking}).

\begin{table}[t]
\centering
\small
\begin{tabular}{lp{8.6cm}}
\toprule
Hypothesis & Verdict \\
\midrule
Answer-space guessing & Rejected: 48.9\% accuracy against an 8.7\%
aggregate guessing ceiling, with a diverse answer distribution. \\
Format-reward hacking & Rejected: correct rows carry genuine multi-step
reasoning (median 270 characters). \\
Template leakage from training & Partial: roughly 3 pp of attributable
inflation; rows without a template duplicate in the training pool still
score 45.9\% against a 3.5\% base. \\
Composition artifacts on \oldset{} & Confirmed---and followed to the
evaluation-validity finding of Section~\ref{sec:results}. \\
\bottomrule
\end{tabular}
\caption{Four reward-hacking hypotheses and their verdicts.}
\label{tab:hacking}
\end{table}

\subsection{Hard gates before training}
\label{sec:gates}

Every load-bearing mechanism was gated by a test that had to pass before
any training step ran: file-level MD5 identity of the teacher and student
tokenizers; a critic reachability gate confirming that image features
influence the value head (identical token sequences with different pixels
must produce different values---observed difference 1.93); a
geometric-ratio self-test for ReOPD's cut-index sampler; the hint-unlock
measurement of Section~\ref{sec:pool} before committing to the pool; and a
projection test showing that HL-Gauss targets are normalized and
mean-preserving.

\subsection{Engineering discipline}
\label{sec:engineering}

All framework modifications are environment-gated patches with self-tests
and \texttt{.orig} backups: with the gating variables unset, the framework's
behavior is bit-identical to stock. Round~4 added three such patches
(hybrid distillation gating, decoupled GAE, and the HL-Gauss critic).

The gates also caught failures in flight rather than in post-mortem:
dynamically injected hints bypassed the framework's overlong-prompt filter
until a 2{,}062-token sample killed all three hint arms at the same step;
the 7B teacher's unprompted answers proved too short to replay (median 9
tokens) and had to be re-sampled with a stored lead-in phrase; and swapping
in the categorical value head required replacing an inner submodule of the
framework's composite value-head wrapper---an error caught by a smoke test,
not by reading code.

\section{Limitations}
\label{sec:limitations}

\paragraph{Single training seed.}
All eleven arms were trained with seed 42. Round-3 measurements of the same
GRPO recipe across three seeds put the seed-to-seed standard deviation at
roughly 2.4--2.7 pp (the range reflects two scoring pipelines). The headline
between-group gaps ($+8$ to $+14$ pp in-domain) clear that noise
comfortably, but differences under about 5 pp---including the entire
ordering \emph{within} the delivered-prior group---should not be ranked.

\paragraph{The pooled in-domain metric is a summary statistic.}
The 71.5/28.5 weighting of \sparseset{} and \oldset{} reflects the sizes we
happened to have, not a sampling design; the two slices are not random draws
from one population, so the pooled number is a weighted summary rather than
an unbiased in-domain accuracy. Equal weighting would lower the metric's
predictive validity ($\rho = +0.40$ vs.\ $+0.76$;
Section~\ref{sec:results-inversion}). The cleanest single in-domain number
is \sparseset{} alone.

\paragraph{\oldset{} is retained only for comparability.}
On the evidence of Section~\ref{sec:results-inversion} it should not be read
as a generalization measure; it appears in our tables solely to keep them
comparable with rounds 1--3 of this project.

\paragraph{\sparseset{} shares its distribution with training.}
The held-out set is drawn from the same pool and the same template families
as the training data. Template-duplicate inflation is quantified at roughly
3 pp for the strongest arm (non-duplicate rows still score 45.9\% against a
3.5\% base); the load-bearing evidence for transfer is DynaMath, which is
programmatically rendered and shares no items with MathV360K.

\paragraph{The predictivity of \sparseset{} is an observation, not a rule.}
That a held-out slice of the training distribution predicts cross-domain
transfer is not guaranteed in advance---a same-distribution set could
equally have favored arms that overfit the training distribution. In this
setting it did not: the arms that gained most on \sparseset{} also
transferred best. We accordingly draw the negative lesson (a drifted
held-out slice can invert method rankings) as general, and the positive one
(the hard in-domain tail was the best available proxy for transfer) as
specific to this setting.

\paragraph{DynaMath-hard is underpowered.}
With n = 104, even the best arm's margin over base (\arm{saf}, 18.3\%) only
reaches $p = 0.09$; we draw no per-arm conclusions from the hard split
alone.

\paragraph{The comparison spans supervision regimes.}
The distillation arms never see ground-truth answers, while the
reward-driven arms do. This is a property of the methods rather than an
oversight, but comparisons between the two families should be read with
that asymmetry in mind.

\paragraph{Judge variance.}
Repeated blind-judging passes over identical outputs vary by about
$\pm 0.5$ pp (\arm{grpo}: 40.5 / 41.1 / 41.0 across three passes); we treat
differences at that scale as scoring noise.

\paragraph{Future work.}
The immediate next step is multi-seed replication of the group-level
result. Three corrective arms are already in flight: an off-context
importance correction for hint-conditioned rollouts
\citep{agrawal2026offcontext}, and variants that remove advantage
std-normalization, which can pathologically amplify dense terms in
near-all-wrong groups \citep{liu2025understanding,wang2026dark}. Beyond
that, we see the most leverage in making the value of a hint prefix itself
measurable---branching student rollouts from teacher trajectories at graded
depths---which would unify the text and distribution priors under one
credit-assignment scheme.

\section{Conclusion}
\label{sec:conclusion}

On a pool of visual-math problems where a 2B vision--language model almost
never succeeds and GRPO receives gradient from fewer than one rollout group
in six, we compared eleven training arms spanning every prior-injection
family proposed for this regime. The comparison localizes what matters. A
prior helps exactly when it is delivered: hint-guided exploration, an
annealed global teacher-KL, and a well-parameterized value critic all clear
the no-prior baseline cross-domain, while pure imitation of a weak teacher,
failure-gated distillation, and a critic trained by scalar regression do
not. Within the families, the active ingredients are narrower than their
papers suggest: the hint arms' gains survive removing UFT's auxiliary
objective, and a one-line change of value-loss parameterization moves
in-domain accuracy by $+14.4$ pp and flips the arm across the group
boundary.

The finding we most expect to outlive the specific methods, however, is the
evaluation one. A held-out set that had served as this project's
generalization check for four rounds ranks the eleven arms in reverse order
of their true cross-domain transfer---not through noise, but through
composition: a near-chance multiple-choice majority rewards arms for
leaving the base model's guessing behavior intact. Any project that tracks
generalization on a fixed held-out slice is exposed to this failure mode,
and it only became visible here once enough arms existed to correlate the
evaluation sets against a genuinely out-of-domain measure. We suggest that
check---made cheap by any multi-arm comparison---as standing practice, and
we document the verification protocol (blind judging, scoring audits,
pre-training gates, hacking analyses) that kept the numbers in this paper
attached to reality. Multi-seed replication of the group-level result and a
per-item decomposition of the inverted set are the immediate next steps.

\bibliography{refs}

\end{document}